\documentclass[10pt,twocolumn,letterpaper]{article}

\usepackage{wacv}              

\usepackage{graphicx}
\usepackage{amsmath}
\usepackage{amssymb}
\usepackage{booktabs}

\usepackage[table,xcdraw]{xcolor}

\usepackage{tikz} 

\definecolor{darkblue}{HTML}{00429d}
\newcommand{\circlednum}[1]{%
\tikz[baseline=(char.base)]{%
    \node[shape=circle,fill=black,text=white, inner sep=0.5pt] (char) {#1};%
    }%
}

\usepackage[pagebackref,breaklinks,colorlinks]{hyperref}

\newcommand{\truong}[1]{\textcolor{black}{#1}}

\usepackage[capitalize]{cleveref}
\crefname{section}{Sec.}{Secs.}
\Crefname{section}{Section}{Sections}
\Crefname{table}{Table}{Tables}
\crefname{table}{Tab.}{Tabs.}

\def\wacvPaperID{1034} 
\def\confName{WACV}
\def\confYear{2025}

\begin{document}

\title{ENAF: A Multi-\underline{E}xit \underline{N}etwork with an \underline{A}daptive Patch \underline{F}usion \\ for Large Image Super Resolution}

\author{Manh Duong Nguyen\\
Hanoi University of Science and Technology\\
Hanoi, Vietnam\\
{\tt\small duong.nm210243@sis.hust.edu.vn}
\and
Tuan Nghia Nguyen\\
Seoul National University\\
South Korea, Seoul 08826\\
{\tt\small nghiant@capp.snu.ac.kr}
\and
Xuan Truong Nguyen\\
Seoul National University\\
South Korea, Seoul 08826\\
{\tt\small truongnx@capp.snu.ac.kr}
}
\maketitle

\begin{abstract}

   {To accelerate single image super-resolution (SISR) networks on large images (2K-8K), many recent approaches decompose an image into small patches and dynamically determine an execution path according to its difficulty (referred to as a dynamic network). To quantify the hardness of a patch, they mainly rely on a handcrafted assessment score, e.g., edge, which weakly associates a patch's texture with the computational complexity of a SISR model. To address the problem, we introduce ENAF - a dynamic network for SISR with an adaptive patch fusion. Built on top of a backbone, ENAF incorporates multiple early exits (EEs) to tackle the over-parameterized SISR model. More importantly, ENAF plugs a tiny network that estimates PSNR to associate data texture with a computation cost at an EE. Based on the scores, ENAF effectively assigns image patches to an exit, enhancing the quality-complexity trade-off. Extensive experiments on common datasets with popular SISR backbones demonstrate the effectiveness of ENAF in various settings. The source code will be available.}
\end{abstract}


\begin{figure*}[t]
  \centering
   \includegraphics[width=0.90\linewidth]{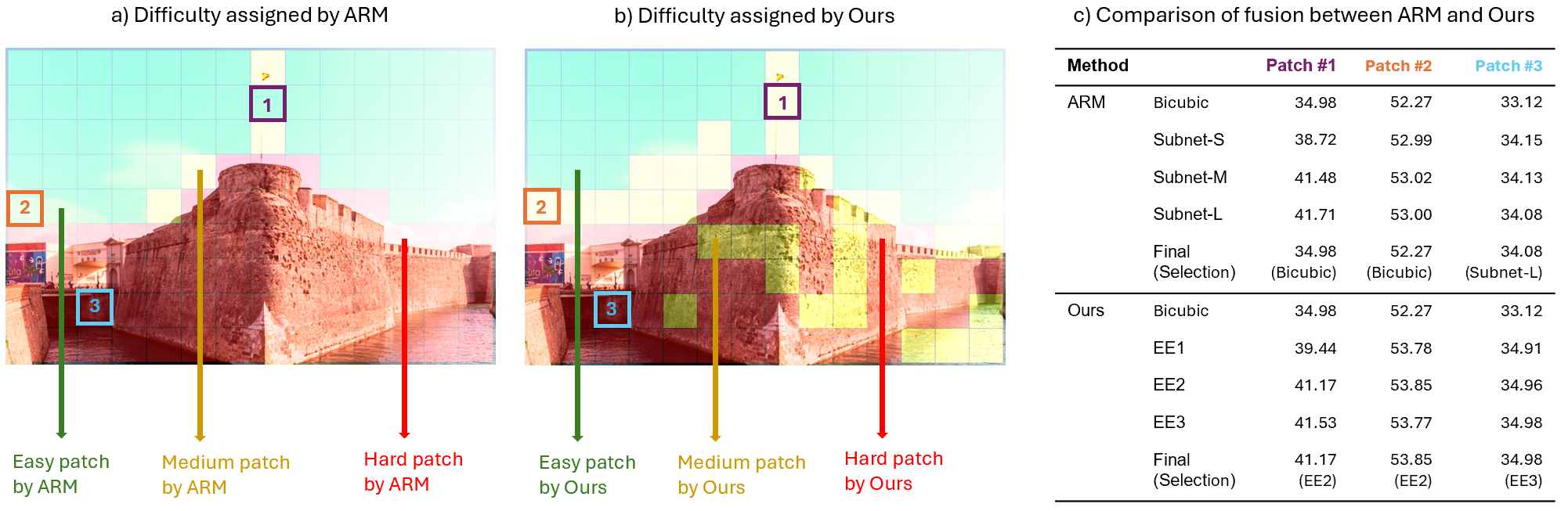}

   \caption{{A visual comparison in patch assignment by ARM \cite{arm2022} and our ENAF. Both reduce the computation cost by approximately 46\%. The portions of "Easy", "Medium", and "Hard" by ARM are 45\%, 3\%, and 52\%, respectively, and that of our ENAF are 38\%, 25\%, and 37\%. With a hand-crafted PSNR-edge-score matching, ARM may fail to classify two patches with little details, i.e., patches 1 and 2. It also inflexibly assigns a patch, i.e., patch 3 to a complex network, thus influencing the performance.}}
   \label{fig:motivation_from_arm}
\end{figure*}

\section{Introduction}
\label{sec:intro}


{Single image super-resolution (SISR) is a long-studied task that refers to constructing a high-resolution (HR) image from a given low-resolution (LR) one \cite{srcnn2014, luo2021funcNN, luo2023AND, lin2024FastSR_NeRF, marcos2024BSRAW}. Since the image/video resolution for display devices has been reaching 2K (1920$\times$1080), 4K (3840$\times$2160), or even 8K (7680$\times$4320), accelerating SISR algorithms on "large" input images has become essential for many practical applications and services such as smartphones and IP TVs. Based on convolutional neural networks (CNNs), the most recent SISR algorithms \cite{srcnn2014, vdsr2016, edsr2017, rdn2018, rcan2018, classsr2021, arm2022}, however, come with a huge memory footprint and a massive computation cost that grow quadratically with the input size \cite{classsr2021, arm2022, wang2024camixersr}.}



{To reduce memory requirements in SISR with a large image, recent SR algorithms decompose input into tiles or patches and execute them one by one \cite{classsr2021, arm2022, wang2024camixersr}. As patches typically overlap with others, i.e., \{tile, overlap\} = \{128,8\}, \{64, 4\}, or \{32,2\} \cite{arm2022, wang2024camixersr}, patch-based execution incurs an overhead in computation or floating-point operations (FLOPs). To reduce computation, \cite{classsr2021, arm2022} have observed that \emph{different image patches have different restoration difficulties and can be processed by networks with different capacities}. ClassSR \cite{classsr2021} incorporated and trained an external classification network (Class-Module) to classify a patch into "easy", "medium", and "hard". ARM \cite{arm2022} introduced a LUT-based classifier and a parameter-sharing design to further improve efficiency. APE \cite{wang2022APE} trained a regressor to predict the incremental capacity of each layer for the patch. Although these approaches are generic to various networks, they suffer two drawbacks: inaccurate patch classification and poor sub-network selection. First, as depicted in Figure 1, a handcrafted classifier in ARM \cite{arm2022} may categorize two low-texture patches in a sample class, i.e., "easy", despite their large differences in restoration performance. Second, sub-network selection rarely incorporates a quality-complexity trade-off among different patches, thus influencing the performance.}

{To address the above shortcomings, we introduce ENAF\footnote{The source code can be found in \url{https://github.com/nmduonggg/ENAF.git}} - a dynamic network for SISR with an adaptive patch fusion. Built on top of a backbone, ENAF incorporates multiple early exits (EEs) to tackle the over-parameterized SISR model. ENAF, however, introduces three new lightweight components. First, unlike \cite{arm2022}, ENAF plugs a tiny \emph{PSNR estimator} that accurately predicts PSNR to associate data texture with a computation cost at an EE. Second, we introduce a simple yet effective blank vector generator to further target patches with little details that account for nearly 13\% in large input images. Third, based on patch scores from the PSNR estimator and the generator, we train an adaptive patch fusion, which makes a good quality-complexity trade-off among different patches. Experiments show that ENAF can help most existing methods (e.g., FSRCNN \cite{fsrcnn2016}, CARN \cite{carn2018}, SRResNet \cite{srresnet2017}) save 24\%-55\%, 43\%-52\%, 43\%-48\%, and 31\%-52\% FLOPs on F2K, Test2K, Test4K, and Test8K datasets, respectively, which largely outperform other SOTA works such as ClassSR\cite{classsr2021} and ARM \cite{arm2022}.}

\section{Background and Motivations}

\begin{figure*}[t]
  \centering
   \includegraphics[width=0.9\linewidth]{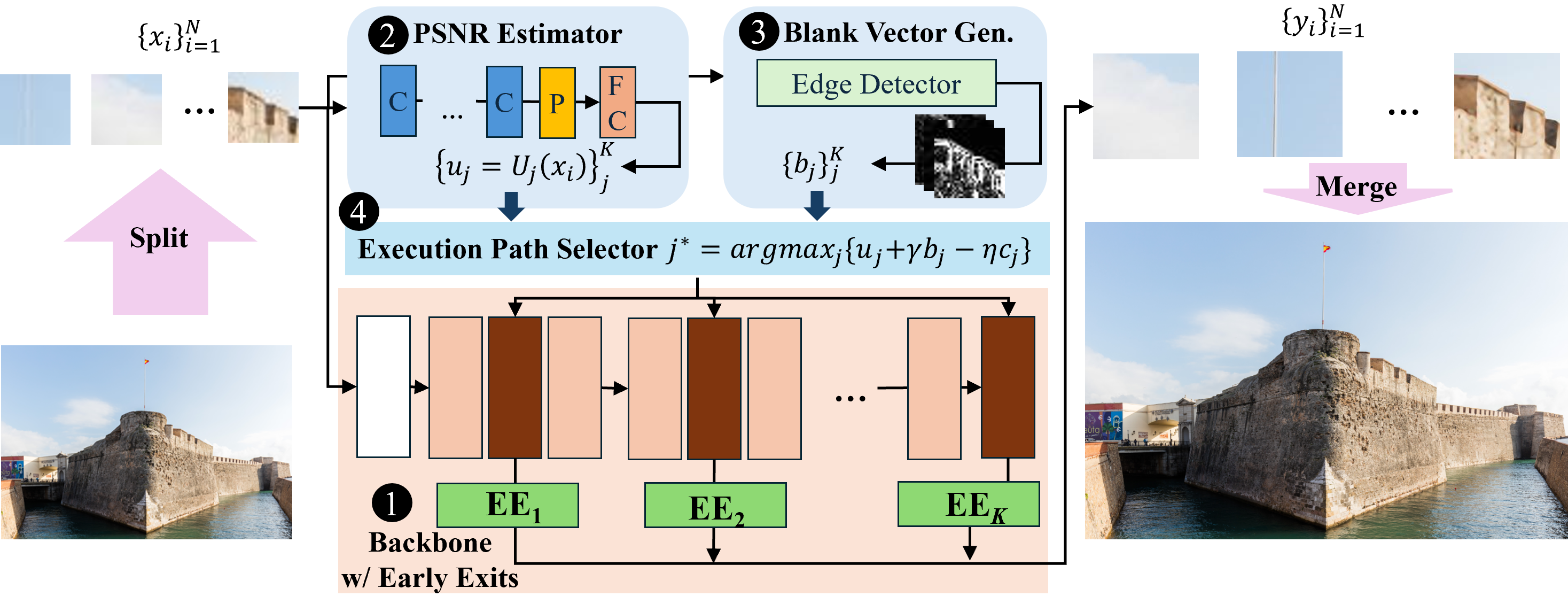}
   \caption{{The Proposed ENAF Architecture.}}
   \label{fig:overview_method}
\end{figure*}

\subsection{Static Super-Resolution Networks and Challenges for Large Input Upscaling} 
{Under the popularity of deep learning in recent years, most SR algorithms have been built on convolutional neural networks (CNNs). SRCNN \cite{srcnn2014} is the first work to build a three-layer convolutional network for high-resolution image reconstruction. VDSR \cite{vdsr2016}, SRResnet \cite{srresnet2017}, and SRDenset \cite{srrdensenet2017} have improved the performance of the SISR task via increase in networks' depth or skip connection integration, such as residual blocks and dense blocks. EDSR \cite{edsr2017} reveals that the batch normalization (BN) layer destroys the scale information of the input images, which is sensitive to super-resolution. Thus, it promotes the elimination of BN layers in a SISR task.}

{High-resolution (HR) images and videos are essential for mobile applications to offer better user experiences, which demands cost-effective SR algorithms. Notably, the memory and computational cost of CNN networks will grow quadratically with the input size. An SR-CNN, with a large input size and large feature maps, requires ×1.5 times larger computation workload and ×10 times larger external memory bandwidth than a classification network \cite{SRNPU}. The increasing model complexity also arouses the community to design lightweight SISR models \cite{fsrcnn2016, carn2018, wdsr2018 }. TPSR \cite{tpsr2020} builds a network with network architecture search (NAS). Zhan et al. \cite{nassr2021} combined NAS with parameter pruning to obtain a real-time SR model for mobile devices. These \emph{static networks} are typically trained and applied uniformly across samples. For example, an easy sample might require less computational cost than a difficult one, suggesting \emph{computational redundancy} and \emph{potential computational savings} remain unexploited during inference.}

\subsection{Dynamic Execution and Patch-Based SR Algorithms} 
{\emph{Dynamic networks}, as opposed to static ones, can adaptively select an execution patch according to an instance's difficulty during inference, which makes them be exploited widely in super-resolution tasks, especially for large image inputs, \cite{classsr2021, arm2022, wang2024camixersr}. A large image is usually decomposed into small patches, and patches have different restoration difficulties and can be processed by networks with different capacities. ClassSR \cite{classsr2021} pre-classified the image patches into three groups of difficulty in high-resolution restoration: "easy", "medium", and "hard". These image patches are respectively sent to various independently trained backbone networks with different widths to perform super-resolution. ARM \cite{arm2022} introduces the definition of supernet and subnet with weight-sharing characteristics. It adopts channel pruning to reduce the number of parameters in the original model for various hardware resources, and leverages PSNR-Edge correlation to offer a good computation-performance trade-off during inference.} 

{Several works \cite{smsr2021, fad2021} have realized the redundancy of input images on fined-grain spatial regions in the SISR task. SMSR \cite{smsr2021} introduces sparse masks in spatial and channel dimensions within each encoder block and then performs sparse convolution to reduce the computational burden at the hardware level. FAD \cite{fad2021} discovers that most spatial information is converted by high-frequency signals in frequency space and offers a computationally intensive branch. It adds multiple convolutional branches of different sizes, and each feature map region is fed to one branch w.r.t. its frequency. However, patch-based SR algorithms are generally more hardware-friendly \cite{classsr2021, arm2022} since they may be directly executed by a GPU or a generic neural processing unit (NPU) \cite{SRNPU}.}

\section{Methodology}
\subsection{{ENAF Architecture}}
Inspired by the recent trendy works \cite{SRNPU, classsr2021, arm2022}, ENAF adopts a dynamic patch-based SR structure as shown in Figure \ref{fig:overview_method}. Given a low-resolution (LR) image, ENAF splits it into  $N$ patches $\{x_i\}_{i=1}^N$ of a fixed size, i.e., 32$\times$32. Following common SISR settings, a patch $x_i$ is fed into a network $F(.)$ with parameters $\theta$ to generate its corresponding high-resolution (HR) patch $y_i$, expressed as follows:
\begin{equation}    \label{eq: conventional mean predict}
    y_i = F(\theta, x_i)
\end{equation}
Based on generated HR patches $\{y_i\}_{i=1}^N$, ENAF merges them to generate the final HR image. Stacking multiple CNN layers or blocks, a SR network $F(.)$ typically consists of feature extraction layers or $L$ encoders $e^{1}, e^{2}, \dots, e^{L}$ followed by sub-pixel or deconvolution layers $f^{L}$. The prediction in Eq. \ref{eq: conventional mean predict} hence can be rewritten as: 
\begin{equation}    \label{eq: original prediction flow}
    \hat{y}_i = ( f^L \prod_{j=1}^L e^j ) (x_i)
\end{equation}
Motivated by the observation that different patches may require different computational capacities, the proposed ENAF incorporates the four main blocks: a backbone network with multiple early exits (\protect\circlednum{1}), a PSNR estimator (\protect\circlednum{2}), a blank vector generator (\protect\circlednum{3}), and an adaptive execution patch selection or fusion (\protect\circlednum{4}), as shown in Figure \ref{fig:overview_method}.

\subsubsection{Backbone with Multi Early Exits}
Inspired by \cite{ape2022}, we modify the original SR backbones to multi-exit networks and present a training strategy in \cref{method: training multi-exit}. However, instead of sharing the same sub-pixel layer $f^L$ as in \cite{ape2022}, ENAF places $K-1$ additional learnable mappings, namely early exits properly along the inference graph of SR network, which have indices $j \in [0, L)$. For simplicity, we define the original predictor of the SR backbone as $K$-th learnable mapping with index $L$, and bicubic operation as the $0$-th execution path. We refer to the index set of all exits in the modified network as $K \subset [0, L]$. Therefore, outcomes from the (early) exit with index $j \in K$, including the final one, can be described as:
\begin{equation}    \label{eq: intermediate prediction flow}
    \hat{y}^j_i = (f^j \prod_{k=1}^{j} e^k ) (x_i)
\end{equation}
where $y_i$ represents the HR image, $f^j$ is the deconvolution layer placed after the intermediate representation generated by $j$-th encoder of the network body.

\subsubsection{{PSNR Estimator}}
\label{method: psnr estimator} 
In order to make the multi-exit SR networks scalable, we propose to design PSNR estimator $U_j$ as a generator that produces an early-exit signal at $j$-th early exit given a sub-image. Similar to \cite{classsr2021} that adopts a difficulty classifier, the PSNR estimator primarily aims to answer \emph{whether the input sub-image is difficult or easy for restoration}. However, \cite{classsr2021} strictly assigns an image patch with a group of hardness, which may reduce the flexibility of patch fusion. Instead, we formulate the hardness evaluation problem under the regression task, precisely the PSNR estimation problem. Despite sharing the same architecture as \cite{classsr2021}, we design this component to generate a PSNR vector instead of a probability vector. This lightweight network ensures no significant computation burden contributed to the main SR component.

\subsection{ENAF Training}
\subsubsection{Training a Multi-Exit SR Network} \label{method: training multi-exit}

For training multiple exits with a similar function, a simple $L_1$ loss function can be used for training all exits jointly as follows:
\begin{equation} \label{eq: all predictors loss}
    \mathcal{L} = \frac{1}{N} \sum_{i=1}^N \sum_{j\in K} |\hat{y}_i^j - y_i|
\end{equation}
The optimization problem in equation \cref{eq: all predictors loss} assigns equal importance to $K$ predictions, including the final output of the original network, which may introduce inconsistencies. 
To clarify, the objective is to learn a model where all predictors perform well simultaneously, achieving similar performance across them. However, this approach does not align with our primary goal of cost reduction without compromising performance. Instead, we split equation \cref{eq: all predictors loss} into two phases, aligning with \cref{eq: main backbone loss,eq: multi-exit loss}.

First, we train the main backbone in a conventional manner, which adopts a simple $L_1$ loss function as follows:
\begin{equation}    \label{eq: main backbone loss}
    \mathcal{L}_{main} = \frac{1}{N} \sum_{i=1}^N |\hat{y}_i^L - y_i|
\end{equation}
This pre-training phase enables us to predefine prior general knowledge on the original weights and guarantee performance consistency compared to the original backbone. 

In the second stage of training, we focus on developing early-exit predictors using knowledge from the initial stage. Since the network is already optimized for the last exit, it is deprioritized in this step. The multi-exit loss, as defined in \cref{eq: multi-exit loss}, trains early-exit predictors while preserving previously learned weights:
\begin{equation} \label{eq: multi-exit loss}
 \mathcal{L}_{ME} = \frac{1}{N} \sum_{i=1}^N \sum_{j \in K} |\hat{y}_i^j - y_i|
\end{equation}
Unlike the first phase, the backbone parameters are now trained with a small learning rate (~$10^{-8}$) to prevent distortion of prior knowledge and avoid performance degradation, which corresponds to a smaller loss weight from the last exit.

\subsubsection{Training a PSNR Estimator}
After a two-stage training process in Sec.\ref{method: training multi-exit}, we add a PSNR estimator and freeze the parameters of the super-resolution backbone. The PSNR estimator then tries to predict the PSNR metric value of each outcome generated by different exits via a hardness evaluation objective loss:
\begin{equation}
 \mathcal{L}_{HE} = \frac{1}{N} \sum_{i=1}^N \sum_{j \in K} |U_{j}(x_i) - P(\hat{y}_i^j, y_i)|
\end{equation}
where $P(a, b)$ refers to function computing PSNR value of image pair ($a, b$).

\subsection{{ENAF Inference and Execution Path Selection}}
In short, we first split a given large LR image into some local sub-images of the same size and perform the PSNR estimation on them. For each exit, we further pre-calculate its computation cost and propose choosing the exit for SR inference with expected better PSNR performance but a smaller computation cost to achieve a computation-performance tradeoff. To produce the final image, all SR patches are combined into a complete SR image.
\subsubsection{Blank Support Vector} \label{sec: blank patch}
\begin{figure}[t]
  \centering
   \includegraphics[width=0.8\linewidth]{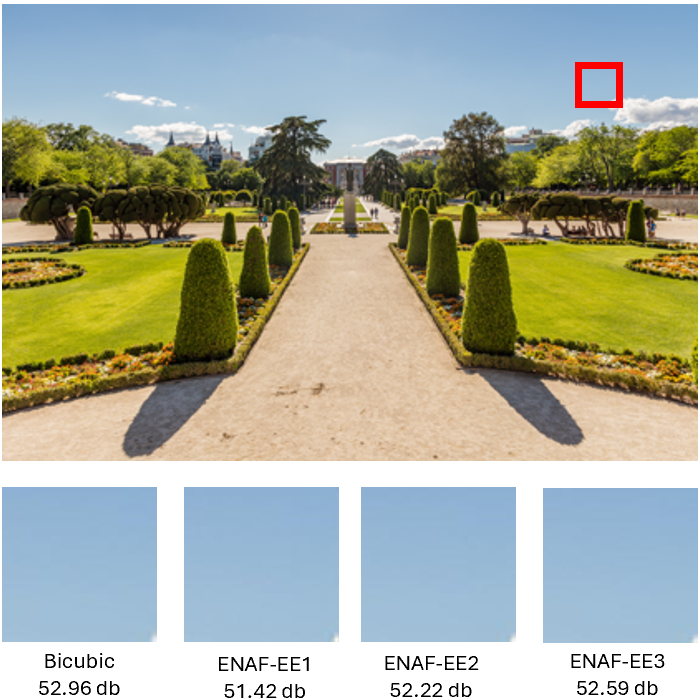}
   \caption{A visual example in which the Bicubic operation performs better than all exits of the SR backbone.}
   \label{fig:blank_patches_why}
\end{figure}

We observed that PSNR estimation for the smallest exit sometimes favors incorrect patterns, as the PSNR Estimator likely assumes smaller networks will perform worse. While reasonable in most cases, this assumption may degrade performance in real-time inference. As shown in \cref{fig:blank_patches_why}, simple bicubic operations can better reconstruct blank patches, i.e., those with minimal texture, than complex neural networks, with minimal computational cost. This is because these patches, representing a small portion of the training data, contribute little to the objective function.

This observation inspires a straightforward idea to detect image blankness and guide it to a simple exit, such as a bicubic operation to improve performance with less computation needed. Since blankness is defined by low texture density, we apply the Laplacian edge detection filter \cite{laplacian1980}, as described in \cite{arm2022} to detect sudden changes in gray intensities. A blank patch, with shortage of texture, should have low pixel values. To capture the overall edge information within the patch, we use the mean value of all pixels in the edge-detected version as the edge score $\hat{p_i}$. A predefined threshold will determine whether a patch is blank. Mathematically, a blank vector is defined as follows:
\begin{equation}    \label{eq:blank vector}
    B(x_i) = 
    \begin{cases}
        [1 \ 0 \ 0 \ ... \ 0], & \text{if } \hat{p_i} \leq \epsilon\\
        [0 \ 0 \ 0 \ ... \ 0], & \text{otherwise}
    \end{cases}
\end{equation}
where $\epsilon$ is a blank threshold (10 in this work). The vector is designed to have a non-zero value only in the first index to primarily prevent bias and incorrect patterns in the PSNR estimator for the simple branch. Therefore, this vector function ensures that only the potentially biased predictor is adjusted.
\begin{figure*}[t]
  \centering
   \includegraphics[width=0.88\linewidth]{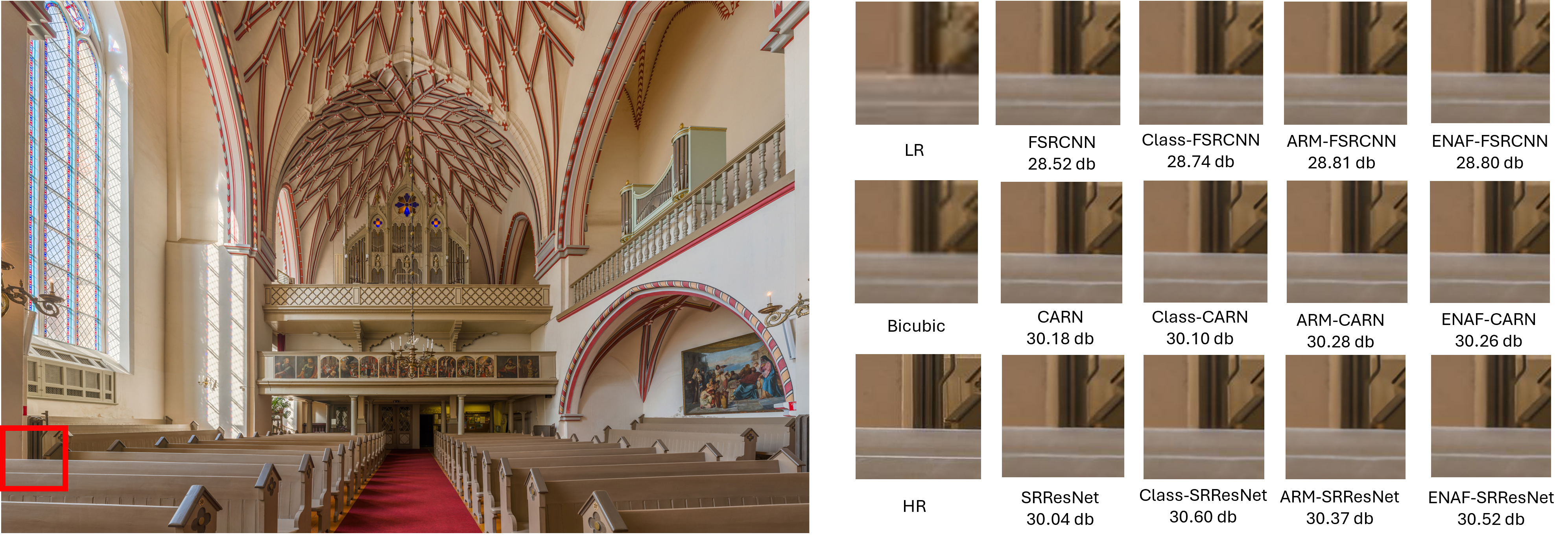}
   \includegraphics[width=0.88\linewidth]{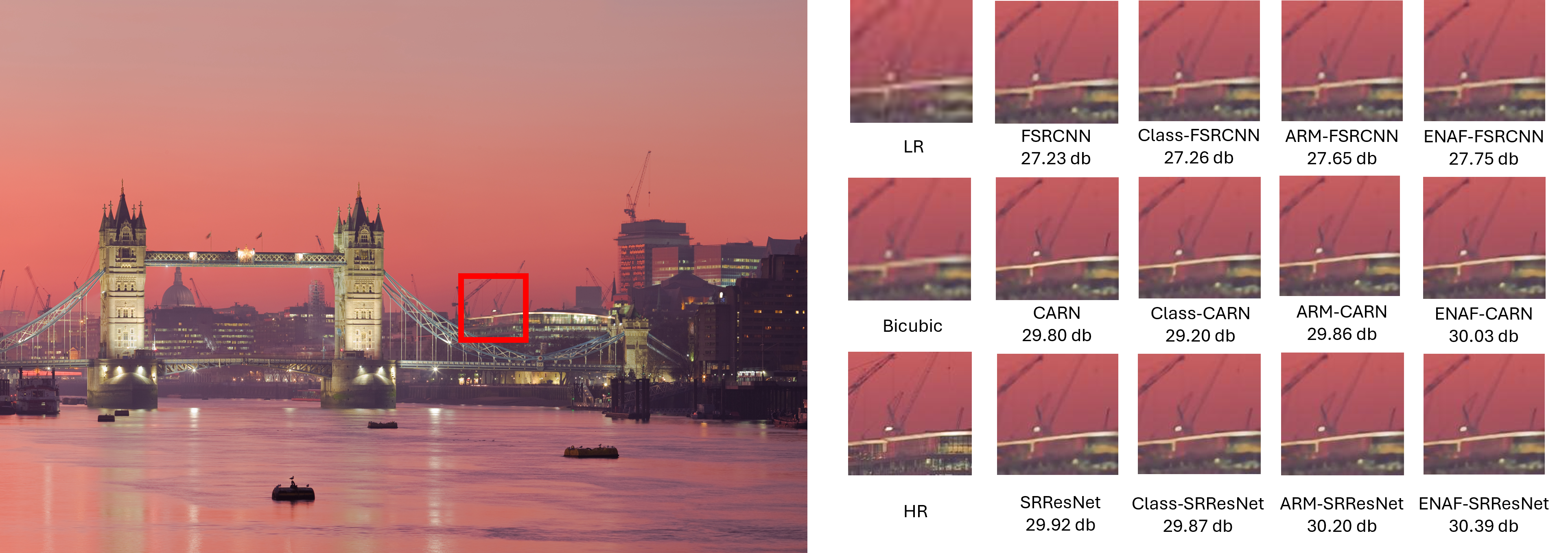}
   \caption{
   {Quantitative comparison of main backbones of ENAF and SOTA patch-based dynamic SISR approaches \cite{classsr2021, arm2022} in $\times 4$ super-resolution settings. These two examples are image "1228" (above) from Test2K and image "1321" (below) from Test4K, respectively. ENAF's main backbone produces an equivalent or higher PSNR compared to original and SOTA methods.}
   }
   \label{fig:better_training_test4k}
\end{figure*}
\subsubsection{Computation-Aware Fusion} 
Given that a deeper exit may slightly improve performance but require a significant increase in computation, we suggest bringing in the theoretical computational burden of each exit. Additionally, due to the incorrect patterns observed in blank sub-images, we propose integrating the blank vector outlined in \cref{eq:blank vector} into the fusion process.Let \(\{c_j\}^K_{j}\) be a set of computation costs and \(\{u_j = U_j(x_i)\}^K_{j}\) a PSNR set, where \(c_j\) and \(u_j\) represent the normalized cost and estimated PSNR for the \(j\)-th execution path. We propose a computation-aware fusion strategy that incorporates the blank vector \(\{b_j\}_j^K = B(x_i)\) as follows:
\begin{equation} \label{eq: inference decision}
    j^* = \underset{j}{\operatorname{argmax}} \ u_j +  \gamma b_j - \eta c_j 
\end{equation}
where $\gamma$ controls contribution of blank vector, $\eta$ is a tradeoff coefficient between performance and computation. The equation \cref{eq: inference decision} guarantees a preference for the $j$-th execution path that provides a higher PSNR estimation while requiring less computational effort and avoiding incorrect patterns, as explained \cref{sec: blank patch}.

\section{Experiments} \label{sec: experiments}
\begin{table*}[t]
\centering
\caption{The comparison of various methods and our ENAF on different datasets F2K, Test2K, Test4K and Test8K}
\label{tab:main result}
\resizebox{0.95\linewidth}{!}{%
\begin{tabular}{@{}
>{\columncolor[HTML]{FFFFFF}}l 
>{\columncolor[HTML]{FFFFFF}}l 
>{\columncolor[HTML]{FFFFFF}}l 
>{\columncolor[HTML]{FFFFFF}}l 
>{\columncolor[HTML]{FFFFFF}}l 
>{\columncolor[HTML]{FFFFFF}}l 
>{\columncolor[HTML]{FFFFFF}}l 
>{\columncolor[HTML]{FFFFFF}}l 
>{\columncolor[HTML]{FFFFFF}}l @{}}
\toprule
\textbf{Model} &
  \textbf{F2K} &
  \textbf{FLOPs} &
  \textbf{Test2K} &
  \textbf{FLOPs} &
  \textbf{Test4K} &
  \textbf{FLOPs} &
  \textbf{Test8K} &
  \textbf{FLOPs} \\ \midrule
FSRCNN &
  27.91 (+0.00) &
  468M (100\%) &
  25.61 (+0.00) &
  468M (100\%) &
  26.90 (+0.00) &
  468M (100\%) &
  32.66 (+0.00) &
  468M (100\%) \\
ClassSR &
  27.93 (+0.02) &
  297M (63\%) &
  25.61 (+0.00) &
  311M (66\%) &
  26.91 (+0.01) &
  286M (61\%) &
  32.73 (+0.07) &
  238M (51\%) \\
ARM-L &
  27.98 (+0.07) &
  380M (82\%) &
  25.64 (+0.03) &
  366M (78\%) &
  26.93 (+0.03) &
  341M (73\%) &
  32.75 (+0.09) &
  290M (62\%) \\
ARM-M &
  27.91 (+0.00) &
  276M (59\%) &
  25.61 (+0.00) &
  289M (62\%) &
  26.90 (+0.00) &
  282M (64\%) &
  32.73 (+0.07) &
  249M (53\%) \\
ARM-S &
  27.65 (-0.26) &
  184M (39\%) &
  25.59 (-0.02 &
  245M (52\%) &
  26.87 (-0.03) &
  230M (50\%) &
  32.66 (+0.00) &
  187M (40\%) \\ \midrule
ENAF-L &
  27.98 (+0.07) &
  \textbf{362M (77\%)} &
  25.64 (+0.03) &
  \textbf{316M (68\%)} &
  26.93 (+0.03) &
  \textbf{301M (64\%)} &
  32.75 (+0.09) &
  \textbf{207M (44\%)} \\
ENAF-M &
  27.91 (+0.00) &
  \textbf{231M (49\%)} &
  25.61 (+0.00) &
  \textbf{229M (49\%)} &
  26.90 (+0.00) &
  \textbf{222M (47\%)} &
  32.73 (+0.07) &
  \textbf{192M (41\%)} \\
ENAF-S &
  27.65 (-0.26) &
  \textbf{112M (24\%)} &
  25.59 (-0.02) &
  \textbf{201M (43\%)} &
  26.87 (-0.03) &
  \textbf{201M (43\%)} &
  32.66 (+0.00) &
  \textbf{144M (31\%)} \\ \midrule \midrule
CARN &
  28.68 (+0.00) &
  1.15G (100\%) &
  25.95 (+0.00) &
  1.15G (100\%) &
  27.34 (+0.00) &
  1.15G (100\%) &
  33.18 (+0.00) &
  1.15G (100\%) \\
ClassSR &
  28.67 (-0.01) &
  766M (65\%) &
  26.01 (+0.06) &
  841M (71\%) &
  27.42 (+0.08) &
  742M (64\%) &
  33.24 (+0.06) &
  608M (53\%) \\
ARM-L &
  28.76 (+0.08) &
  1046M (89\%) &
  26.04 (+0.09) &
  945M (80\%) &
  27.45 (+0.11) &
  825M (70\%) &
  33.31 (+0.13) &
  784M (66\%) \\
ARM-M &
  28.68 (+0.00) &
  819M (69\%) &
  26.02 (+0.07) &
  831M (71\%) &
  27.42 (+0.08) &
  743M (64\%) &
  33.27 (+0.09) &
  612M (53\%) \\
ARM-S &
  28.57 (-0.11) &
  676M (57\%) &
  25.95 (+0.00) &
  645M (55\%) &
  27.34 (+0.00) &
  593M (50\%) &
  33.18 (+0.00) &
  489M (42\%) \\ \midrule
ENAF-L &
  28.76 (+0.08) &
  \textbf{1012M (86\%)} &
  26.04 (+0.09) &
  \textbf{932M (79\%)} &
  27.45 (+0.11) &
  835M (71\%) &
  33.31 (+0.13) &
  \textbf{617M (52\%)} \\
ENAF-M &
  28.68 (+0.00) &
  \textbf{709M (60\%)} &
  26.02 (+0.07) &
  \textbf{698M (59\%)} &
  27.42 (+0.08) &
  \textbf{648M (55\%)} &
  33.27 (+0.09) &
  \textbf{581M (49\%)} \\
ENAF-S &
  28.57 (-0.11) &
  \textbf{644M (55\%)} &
  25.95 (+0.00) &
  \textbf{610M (52\%)} &
  27.34 (+0.00) &
  \textbf{572M (48\%)} &
  33.18 (+0.00) &
  510M (44\%) \\ \midrule \midrule
SRResNet &
  29.01 (+0.00) &
  5.20G (100\%) &
  26.19 (+0.00) &
  5.20G (100\%) &
  27.65 (+0.00) &
  5.20G (100\%) &
  33.50 (+0.00) &
  5.20G (100\%) \\
ClassSR &
  29.02 (+0.01) &
  3.43G (66\%) &
  26.20 (+0.01) &
  3.62G (70\%) &
  27.66 (+0.01) &
  3.30G (63\%) &
  33.50 (+0.00) &
  2.70G (52\%) \\
ARM-L &
  29.03 (+0.02) &
  4.23G (81\%) &
  26.21 (+0.02) &
  4.00G (77\%) &
  27.66 (+0.01) &
  3.41G (66\%) &
  33.52 (+0.02) &
  3.24G (62\%) \\
ARM-M &
  29.01 (+0.00) &
  3.59G (69\%) &
  26.20 (+0.01) &
  3.48G (67\%) &
  27.65 (+0.01) &
  3.30G (62\%) &
  33.50 (+0.00) &
  2.47G (48\%) \\
ARM-S &
  28.97 (-0.04) &
  2.74G (53\%) &
  26.18 (-0.01) &
  2.87G (55\%) &
  27.63 (-0.02) &
  2.77G (53\%) &
  33.46 (-0.04) &
  1.83G (35\%) \\ \midrule
ENAF-L &
  29.03 (+0.02) &
  \textbf{3.23G (62\%)} &
  26.21 (+0.02) &
  \textbf{3.07G (59\%)} &
  27.66 (+0.01) &
  \textbf{2.81G (54\%)} &
  33.52 (+0.02) &
  \textbf{2.43G (47\%)} \\
ENAF-M &
  29.01 (+0.00) &
  \textbf{3.04G (58\%)} &
  26.20 (+0.01) &
  \textbf{2.82G (54\%)} &
  27.65 (+0.01) &
  \textbf{2.66G (51\%)} &
  33.50 (+0.00) &
  \textbf{2.15G (41\%)} \\
ENAF-S &
  28.97 (-0.04) &
  2.74G (53\%) &
  26.18 (-0.01) &
  \textbf{2.61G (50\%)} &
  27.63 (-0.02) &
  \textbf{2.39G (46\%)} &
  33.46 (-0.04) &
  1.88G (36\%) \\ \bottomrule
\end{tabular}%
}
\end{table*}

\subsection{Settings}
\subsubsection{Training Datasets and Backbones}
\textbf{Datasets and Preprocessing.} Targeting large images, the DIV2K training set (indices 0001-0800) \cite{div2k} is used for all the training phases, similar to \cite{classsr2021, arm2022}. The images in the training set will be pre-processed and decomposed into small patches. During training, data augmentation techniques included random rotation and flipping.

\noindent \textbf{Backbones.} We construct the ENAF framework on three backbones with different network sizes, FSRCNN \cite{fsrcnn2016} (small), CARN \cite{carn2018} (medium), and SRResNet \cite{srresnet2017} (large), following \cite{classsr2021, arm2022}. We configured four exits within each backbone, including Bicubic, EE1, EE2, and EE3, where EE3 is the last exit of the original network. We assign all exits at reasonable positions alongside the forward flow of the network. For more details, the proportions are (31\%, 67\%, 100\%) for FSRCNN, (66\%, 74\%, 100\%) for CARN, and (39\%, 71\%, 100\%) for SRResNet. 
\vspace{-4pt}
\subsubsection{Training Configurations}
Our training strategy has three main phases. First, we pre-train the SR backbone until appropriate convergence. The pre-training phase spans 30 epochs with a batch size of 16. $L_1$ loss function \cite{l12004} is adopted with Adam optimizer \cite{adamoptim2015}  ($\eta1 = 0.9$, $\beta2 = 0.999$). Additionally, a cosine annealing learning strategy is used to adjust the learning rate dynamically along the training period, with a minimum value of $10^{-8}$. The learning rate is initially set to $10^{-3}$ and reduced after each epoch. 
Second, we add two early exits in the predefined positions and finetune the module. While additional early exits are trained with an initial learning rate of $10^{-3}$, the pre-trained parts are finetuned with a small learning rate only ($10^{-8}$ in this work), not to destroy weights learned previously. The batch size in this phase is set to 128 due to increased parameters.
Finally, we train our PSNR estimator with the same setting as the second phase. During this process, the parameters of the backbone network are frozen. All models are built on PyTorch framework \cite{pytorch2017} and trained on NVIDIA RTX-3090 GPUs.

\subsubsection{Benchmark and Evaluation Metric}
\textbf{Evaluation Datasets.} Following \cite{classsr2021, arm2022}, we evaluate all methods on four large-size test images, including Flickr2K, Test2K, Test4K, and Test8K. The Test2K, Test4K, and Test8K datasets are derived from DIV8K \cite{div8k} by downsampling as mentioned in previous work \cite{classsr2021}. 

\noindent \textbf{Metric.} We adopt two common metrics: the Peak Signal-to-Noise Ratio (PSNR) and FLOPs for computation cost. All other evaluation settings adhere to the standard protocols outlined in {\cite{classsr2021, arm2022}, i.e., we fix the PSNR and compare the FLOPS required to achieve that value.}. Unless explicitly stated, the FLOPS mentioned in this study are calculated as the mean FLOPS for all 32×32 low-resolution patches with ×4 super-resolution across the entire test set to ensure a fair comparison with the results presented in \cite{arm2022}. 

\noindent \textbf{Hyperparameters.} ENAF and \cite{arm2022} offer a strategy to achieve computation-performance tradeoff leading to various possible combinations of computation-performance. During inference, two important hyperparameters to tune are $\gamma$ and $\eta$. While $\gamma$ controls the effect of prior knowledge of blankness in decision and is set to 1.0 - 1.5, $\eta$ controls the tradeoff between computation and restoration accuracy, which is dynamically adjusted to generate multiple versions of the multi-exit network.

\begin{table*}[t]
\centering
\caption{Ablation study on effect of blank vector generator}
\label{tab:abl study - remove blank}
\resizebox{0.70\linewidth}{!}{%
\begin{tabular}{@{}
>{\columncolor[HTML]{FFFFFF}}l 
>{\columncolor[HTML]{FFFFFF}}l 
>{\columncolor[HTML]{FFFFFF}}l 
>{\columncolor[HTML]{FFFFFF}}l 
>{\columncolor[HTML]{FFFFFF}}l 
>{\columncolor[HTML]{FFFFFF}}l 
>{\columncolor[HTML]{FFFFFF}}l @{}}
\toprule
\textbf{Model} &
  \textbf{F2K} &
  \textbf{FLOPs} &
  \textbf{Test2K} &
  \textbf{FLOPs} &
  \textbf{Test4K} &
  \textbf{FLOPs} \\ \midrule
ENAF-L-NB &
  \cellcolor[HTML]{FFFFFF}27.98 (+0.07) &
  379M (81\%) &
  \cellcolor[HTML]{FFFFFF}25.64 (+0.03) &
  368M (79\%) &
  \cellcolor[HTML]{FFFFFF}26.93 (+0.03) &
  375M (80\%) \\
ENAF-M-NB &
  \cellcolor[HTML]{FFFFFF}27.91 (+0.00) &
  262M (56\%) &
  \cellcolor[HTML]{FFFFFF}25.61 (+0.00) &
  271M (58\%) &
  \cellcolor[HTML]{FFFFFF}26.90 (+0.00) &
  284M (61\%) \\
ENAF-S-NB &
  \cellcolor[HTML]{FFFFFF}27.65 (-0.26) &
  105M (23\%) &
  \cellcolor[HTML]{FFFFFF}25.59 (-0.02) &
  247M (53\%) &
  \cellcolor[HTML]{FFFFFF}26.87 (-0.03) &
  256M (54\%) \\ \midrule
ENAF-L &
  27.98 (+0.07) &
  \textbf{362M (77\%)} &
  25.64 (+0.03) &
  \textbf{316M (68\%)} &
  26.93 (+0.03) &
  \textbf{301M (64\%)} \\
ENAF-M &
  27.91 (+0.00) &
  \textbf{231M (49\%)} &
  25.61 (+0.00) &
  \textbf{229M (49\%)} &
  26.90 (+0.00) &
  \textbf{222M (47\%)} \\
ENAF-S &
  27.65 (-0.26) &
  112M (24\%) &
  25.59 (-0.02) &
  \textbf{201M (43\%)} &
  26.87 (-0.03) &
  \textbf{201M (43\%)} \\ \bottomrule
\end{tabular}%
}
\end{table*}

\subsection{Results}
\cref{tab:main result} shows the quantitative performance comparison of ENAF framework with other computation-aware SISR methods such as \cite{classsr2021, arm2022} on three conventional backbones: FSRCNN \cite{fsrcnn2016}, CARN \cite{carn2018}, and SRResNet \cite{srresnet2017}. Overall, it can be observed that ENAF consistently outperforms ClassSR and ARM with better computation-performance tradeoff, i.e., requires less cost to achieve the same performance.
{In the large version ("-L"), our method significantly reduces computation compared to ARM by 4-12\% of the original FLOPS for all SISR backbones. This trend holds for the "-M" and "-S" versions across multiple datasets, demonstrating our efficiency and performance-computation tradeoff. With similar restoration quality, our ENAF network uses just 45\%, 52\%, and 47\% of the original computation for FSRCNN, CARN, and SRResNet, respectively.}
We also observed that the reduction of FLOPs is likely to correlate with resolution highly. It seems that models achieve the largest computation reduction on Test8K. Intuitively, a larger image typically includes more low-texture patches, increasing the chances of an efficient execution path being chosen.
In \cref{fig:better_training_test4k}, we provide some visual results inferred by the main backbone trained by the strategy from \cite{classsr2021, arm2022} and ours to show that our training strategy is effective in SISR scenarios. The result indicates that the main backbone trained following our strategy can have comparable or better performance than others by roughly 0.2 dB on average. This is because the image restoration objective loss in ClassSR \cite{classsr2021} is interfered with by the classification loss, while ARM's \cite{arm2022} channel pruning mechanism includes all subnets into a single parameter set, potentially degrading SR performance. ENAF, on the other hand, focuses on learning different mappings built upon a shared set of encoders, thus ensuring parameter sharing without significantly affecting the final performance. This approach also allows ENAF to scale to more exit levels with significantly lower extra costs compared to ClassSR~\cite{classsr2021}.

\begin{figure}[h]
  \centering
    \includegraphics[width=0.9\linewidth]{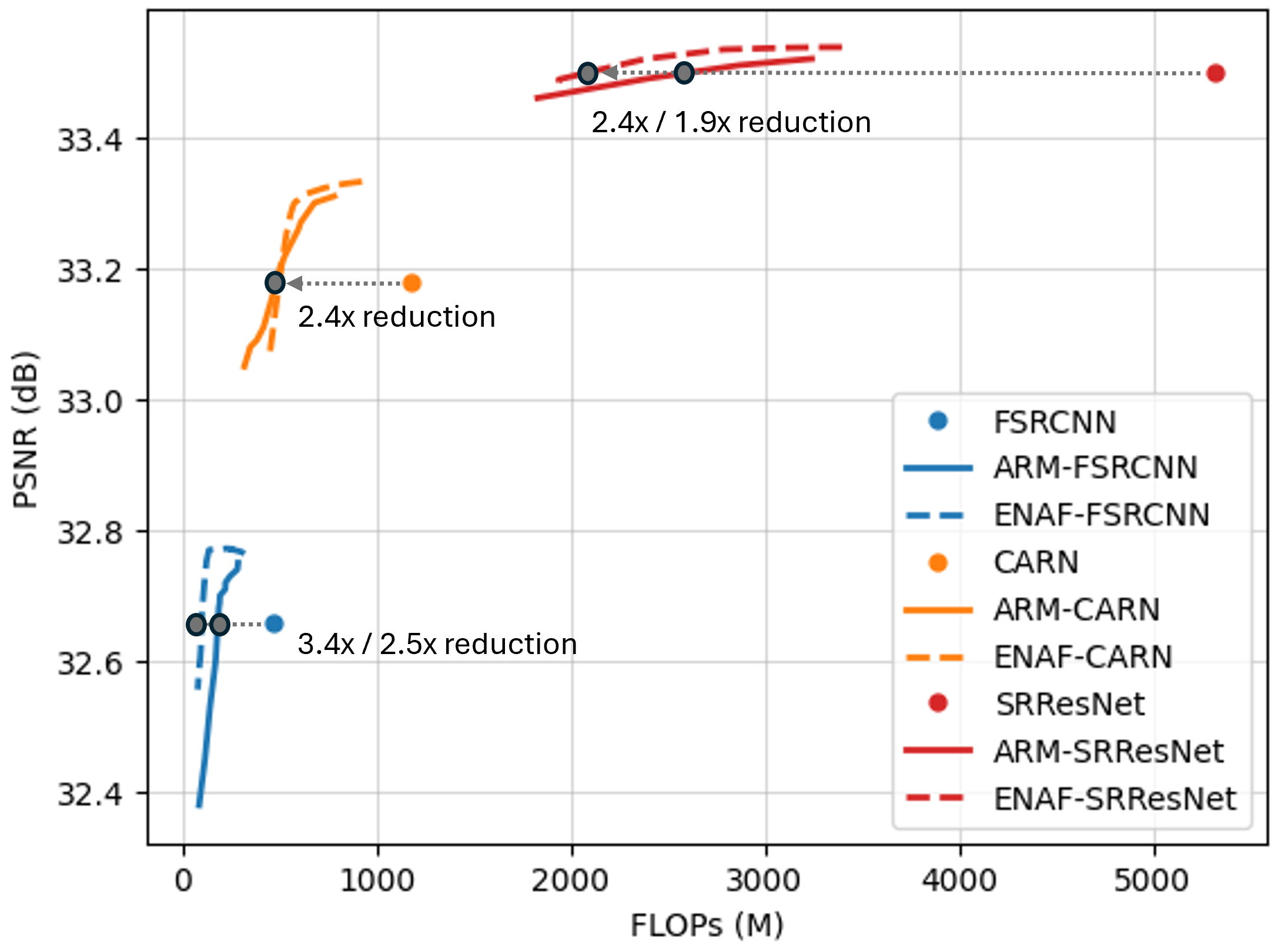}
   \caption{ENAF networks offer less computational requirement than the backbones and ARM to achieve comparable performance, thus better FLOPS-PSNR tradeoffs.}
   \label{fig:performance_range_chart}
\end{figure}

{
Regarding the computation-performance tradeoff, \cref{fig:performance_range_chart} depicts that our ENAF network offers a better or comparable tradeoff to ARM~\cite{arm2022}. The PSNR-FLOPS lines show that ENAF-FSRCNN and ENAF-SRResNet have higher restoration performance and require less computation, indicating that ours has a superior performance-cost tradeoff and a better execution path selection strategy.}
Notably, the superiority observed in FSRCNN is considerable, with an impressive margin. This is because the ability to restore images in lightweight networks like FSRCNN somewhat struggles to outperform a bicubic operation in some simple cases. Thus, the blank vector generated during inference aids its ability to prefer the bicubic approach instead of the learned network, with slight difference in performance.
\subsection{Ablation Study}

\subsubsection{Effect of Blank Vector Generation}
{In the ablation study, we test the effect of the blank vector on helping to reduce computational cost on different datasets with ENAF-FSRCNN. }
Specifically, we remove the blank vector component in the inference strategy, which therefore only takes into consideration performance (PSNR) and computation (FLOPS). We call them "-NB" versions, which stands for "No Blank," to distinguish them from our original proposal. \cref{tab:abl study - remove blank} indicates that without the support of prior knowledge of a given patch's blankness,
ENAF's variants require increases by around 5\% to 15\% of original computation (FLOPS) to reach the similar performance (PSNR). The margin of the cost seems to enlarge, corresponding to the resolution of the images since a larger image can be divided into more sub-images, which allows the blank vector to support biased simple predictors better.

\subsubsection{Effect of PSNR Estimator}
To assess the accuracy of the PSNR estimator in our fusion strategy, we compared the mean absolute error (MAE) between the predicted and actual PSNR values at different depth levels. This comparison was made against ARM \cite{arm2022}, which utilizes an Edge-to-PSNR Lookup Table for patch fusion. To simplify, we designate the inference levels 0, 1, 2, and 3 as Bicubic, Subnet-S, Subnet-M, and Subnet-L when evaluating ARM-LUT. Similarly, we refer to the inference levels as Bicubic, EE1, EE2, and EE3 when using our ENAF-Estimator.
\begin{table}[t]
\centering
\caption{{MAE between the predicted and actual PSNR values at different execution depth levels.}}
\label{tab:effect of estimator}
\resizebox{0.8\linewidth}{!}{%
\begin{tabular}{@{}lrrrr@{}}
\toprule
\textbf{Method} & \textbf{Level 0}     & \textbf{Level 1}     & \textbf{Level 2}     & \textbf{Level 3}     \\ \midrule
ARM LUT         & \textbf{2.50} & 3.41          & 3.47          & 3,48          \\
ENAF Estimator  & 2.93           & \textbf{2.90} & \textbf{2.94} & \textbf{2.93} \\ \bottomrule
\end{tabular}%
}
\end{table}
As illustrated in \cref{tab:effect of estimator}, the ENAF-Estimator outperforms ARM-LUT at three deeper levels with a margin of approximately 0.5 on average. Regarding PSNR estimation of bicubic outcomes, a lookup table performs slightly better (0.43). The experiment is implemented on SRResNet and conducted on the Test2K dataset.

\subsubsection{Relation of $\eta$ and FLOPS}
We also examine the correlation between $\eta$ in fusion strategy as described in \cref{eq: inference decision}.
\begin{figure}[t]
  \centering
   \includegraphics[width=0.85\linewidth]{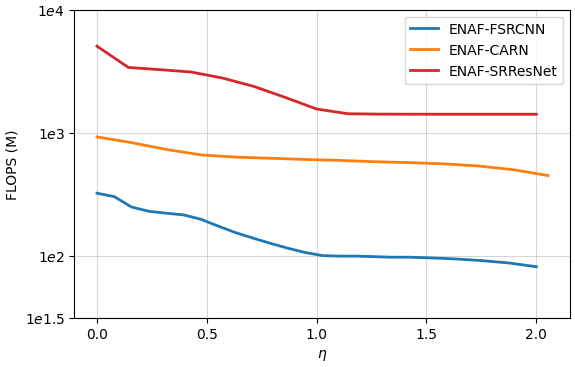}
   \caption{Relation of $\eta$ and FLOPS of three different backbones. Conducted on Test8K dataset}
   \label{fig:eta_flops}
\end{figure}
\cref{fig:eta_flops} indicates that there exits negative correlation between $\eta$ and FLOPS which is obvious since $\eta$ controls the computation-related constraint in fusion strategy. Notably, when $\eta$ ranges from 0 to 1.0, the FLOPS of all three backbones drop considerably with a sharp downward trend. Following this, the FLOPS of all backbones remain nearly constant from $\eta=1.0$ to $\eta=2.0$. This result provides intuition about how $\eta$ and FLOPS interact with each other and, therefore, supports implementing ENAF in real-world applications.

\section{Conclusion}
This study presents an effective dynamic network with multiple early exits for accelerating large image super-resolution.
To adaptively select an execution path during inference, we introduce a simple yet effective strategy based on a tiny PSNR estimator and a blank vector generator. Our design makes a good computation-performance tradeoff across various model scales and computational budgets, making it applicable for practical applications based on efficient super-resolution.

\section{Acknowledgement}
\truong{This research was also supported by the MSIT (Ministry of Science and ICT), Korea, under the ITRC (Information Technology Research Center) support program (IITP-2024-2020-0-01461) supervised by the IITP (Institute for Information \& communications Technology Planning \& Evaluation and in part under the artificial intelligence semiconductor support program to nurture the best talents (IITP-2023-RS-2023-00256081). The corresponding author is Xuan Truong Nguyen.}
{\small
\bibliographystyle{ieee_fullname}
\bibliography{egbib}
}

\end{document}